\documentclass[runningheads]{llncs}
\usepackage{graphicx}

\usepackage{tikz}
\usepackage{comment} 
\usepackage{amsmath,amssymb} 
\usepackage{color}

\usepackage{bm}
\usepackage{booktabs}
\usepackage{siunitx}
\usepackage{stfloats}
\usepackage{multirow}
\usepackage{array}

\begin{document}
\pagestyle{headings}
\mainmatter
\def\ECCVSubNumber{100}  

\title{Dynamic ReLU} 

\titlerunning{Dynamic ReLU}
%
\author{Yinpeng Chen\orcidID{0000-0003-1411-225X} \and
Xiyang Dai\orcidID{0000-0003-1761-8715} \and
Mengchen Liu\orcidID{0000-0002-9040-1013} \and
Dongdong Chen\orcidID{0000-0002-4642-4373} \and
Lu Yuan\orcidID{0000-0001-7879-0389} \and
Zicheng Liu\orcidID{0000-0001-5894-7828}
}
\authorrunning{Chen Y., Dai X., Liu M., Chen D., Yuan L., Liu Z.}
%
\institute{Microsoft Corporation, Redmond WA 98052, USA\\
\email{\{yiche,xidai,mengcliu,dochen,luyuan,zliu\}@microsoft.com}}
\maketitle

\begin{abstract}
Rectified linear units (ReLU) are commonly used in deep neural networks. So far ReLU and its generalizations (non-parametric or parametric) are \textbf{static}, performing identically for all input samples.
In this paper, we propose \textbf{Dynamic ReLU} (DY-ReLU), a dynamic rectifier of which parameters are generated by a hyper function over all input elements.
The key insight is that DY-ReLU \textit{encodes the global context into the hyper function, and adapts the piecewise linear activation function accordingly}.
Compared to its static counterpart, DY-ReLU has negligible extra computational cost, but significantly more representation capability, especially for light-weight neural networks. By simply using DY-ReLU for MobileNetV2, the top-1 accuracy on ImageNet classification is boosted from 72.0\% to 76.2\% with only 5\% additional FLOPs.
\end{abstract}

\section{Introduction}
\begin{figure}[b]
	\begin{center}
		\includegraphics[width=0.76\linewidth]{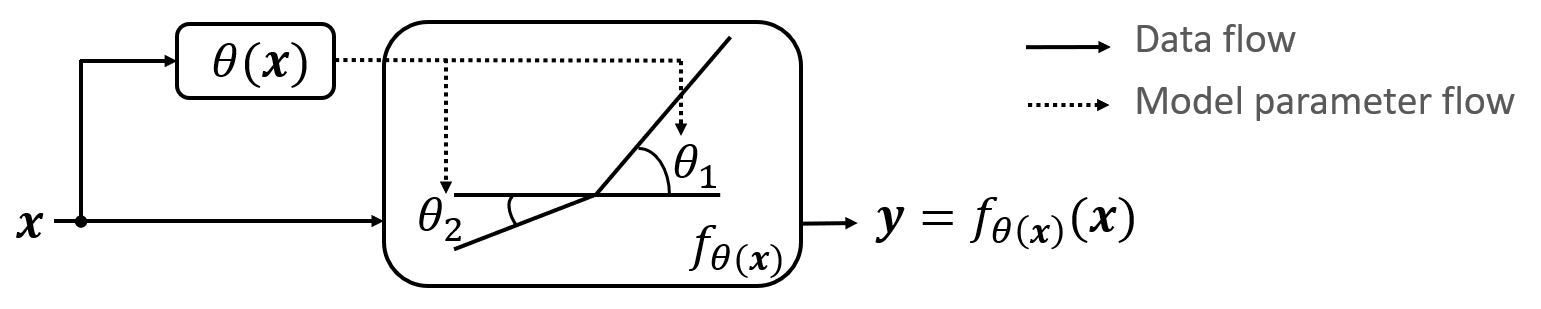}
	\end{center}
	\caption{Dynamic ReLU. The piecewise linear function is determined by the input $\bm{x}$.}
	\label{fig:overview}
\end{figure}

Rectified linear unit (ReLU) \cite{NairH10Relu,JarrettKRL09Relu} is one of the few milestones in the deep learning revolution. It is simple and powerful, greatly improving the performance of feed-forward networks. Thus, it has been widely used in many successful architectures (e.g. ResNet \cite{he2016deep}, MobileNet\cite{howard2017mobilenets,sandler2018mobilenetv2,howard2019mbnetv3} and ShuffleNet \cite{Zhang_2018_CVPR,ma_2018_ECCV}) for different vision tasks (e.g. recognition, detection, segmentation).
ReLU and its generalizations, either non-parametric (leaky ReLU \cite{Maas13rectifiernonlinearities}) or parametric(PReLU \cite{he_2015_prelu}) are \textbf{static}. They perform in the exactly same way for different inputs (e.g. images). This naturally raises an issue: \textit{should rectifiers be fixed or adaptive to input (e.g. images)}? In this paper, we investigate \textit{dynamic} rectifiers to answer this question.  

We propose dynamic ReLU (DY-ReLU), a piecewise function $f_{\bm{\theta}(\bm{x})}(\bm{x})$ whose parameters are computed from a hyper function $\bm{\theta}(\bm{x})$ over input $\bm{x}$. Figure \ref{fig:overview} shows an example that the slopes of two linear functions are determined by the hyper function. The key idea is that the \textit{global context of all input elements} $\bm{x}=\{x_c\}$ is encoded in the hyper function $\bm{\theta}(\bm{x})$ for adapting the activation function $f_{\bm{\theta}(\bm{x})}(\bm{x})$. This enables significantly more representation capability, especially for light-weight neural networks (e.g. MobileNet). Meanwhile, it is computationally efficient as the hyper function $\bm{\theta}(\bm{x})$ is simple with negligible computational cost.

Furthermore, we explore three variations of dynamic ReLU, which share activation functions across spatial locations and channels differently: (a) spatial and channel-shared DY-ReLU-A,  (b) spatial-shared and channel-wise DY-ReLU-B, and (c) spatial and channel-wise DY-ReLU-C. They perform differently at different tasks. Channel-wise variations (DY-ReLU-B and DY-ReLU-C) are more suitable for image classification. When dealing with keypoint detection, DY-ReLU-B and DY-ReLU-C are more suitable for the backbone network while the spatial-wise DY-ReLU-C is more suitable for the head network.  

We demonstrate the effectiveness of DY-ReLU on both ImageNet classification and COCO keypoint detection. Without bells and whistles, simply replacing static ReLU with dynamic ReLU in multiple networks (ResNet, MobileNet V2 and V3) achieves solid improvement with only a slight increase (5\%) of computational cost. 
For instance, when using MobileNetV2, our method gains 4.2\% top-1 accuracy on image classification and 3.5 AP on keypoint detection, respectively.

\section{Related Work}
\noindent \textbf{Activation Functions:} activation function introduces non-linearity in deep neural networks. 
Among various activation functions, ReLU~\cite{hahnloser2000relu2,NairH10Relu,JarrettKRL09Relu} is widely used.
Three generalizations of ReLU are based on using a nonzero slopes $\alpha$ for negative input. Absolute value rectification \cite{JarrettKRL09Relu} fixes $\alpha=-1$. LeakyReLU \cite{Maas13rectifiernonlinearities} fixes $\alpha$ to a small value, while PReLU \cite{he_2015_prelu} treats $\alpha$ as a learnable parameter. 
RReLU took a further step by making the trainable parameter a random number sampled from a uniform distribution~\cite{Xu15empiricalactivaiton}.
Maxout \cite{goodfellow2013maxout} generalizes ReLU further, by dividing input into groups and outputs the maximum. 
One problem of ReLU is that it is not smooth. A number of smooth activation functions have been developed to address this, such as softplus~\cite{dugas2001softplus}, ELU~\cite{clevert2015elu}, SELU~\cite{klambauer2017selu}, Mish~\cite{misra2019mish}. PELU~\cite{trottier2017pelu} introduced three trainable parameters into ELU. 
Recently, empowered by neural architecture search (NAS) techniques \cite{Zoph2017NeuralAS,real2018aaai,Zoph_2018_CVPR,liu2018darts,xie2018snas,cai2018proxylessnas,Tan_2019_CVPR,Wu_2019_CVPR}, Ramachandran et al.~\cite{ramachandran2017searching} found several novel activation functions, such as Swish function. Different with these static activation functions that are input independent, our dynamic ReLU adapts the activation function to the input.

\noindent \textbf{Dynamic Neural Networks:} Our method is related to recent work of dynamic neural networks \cite{NIPS2017_6813,liu2018ddnn,Wang_2018_ECCV,Wu_2018_CVPR,yu2018slimmable,huang2018multiscale,Hu_2018_CVPR,Yang2019CondConvCP,Chen2019DynamicCA}.  D$^2$NN \cite{liu2018ddnn}, SkipNet \cite{Wang_2018_ECCV} and  BlockDrop \cite{Wu_2018_CVPR} learn a controller for skipping part of an existing model by using reinforcement learning. MSDNet \cite{huang2018multiscale} allows early-exit based on the prediction confidence. Slimmable Net \cite{yu2018slimmable} learns a single network executable at different widths. Once-for-all \cite{Cai2019OnceFA} proposes a progressive shrinking algorithm to train one network that supports multiple sub-networks. Hypernetworks \cite{Ha2017HyperNetworks} generates network parameters using anther network. SENet \cite{Hu_2018_CVPR} squeezes global context to reweight channels. Dynamic convolution \cite{Yang2019CondConvCP,Chen2019DynamicCA} adapts convolution kernels based on their attentions that are input dependent. 
Compared with these works, our method shifts the focus from kernel weights to activation functions.

\noindent \textbf{Efficient CNNs:} Recently, designing efficient CNN architectures \cite{squeezenet16,howard2017mobilenets,sandler2018mobilenetv2,howard2019mbnetv3,Zhang_2018_CVPR,ma_2018_ECCV} has been an active research area. MobileNetV1 \cite{howard2017mobilenets} decomposes a $3\times3$ convolution to a depthwise convolution and a pointwise convolution. MobileNetV2 \cite{sandler2018mobilenetv2} introduces inverted residual and linear bottlenecks. MobileNetV3 \cite{howard2019mbnetv3} applies squeeze-and-excitation \cite{Hu_2018_CVPR}, and employs a platform-aware neural architecture search approach \cite{Tan_2019_CVPR} to find the optimal network structure. ShuffleNet further reduces MAdds for $1\times1$ convolution by group convolution. ShiftNet \cite{shift} replaces expensive spatial convolution by the shift operation and pointwise convolution. 
Our method provides an effective activation function, which can be easily used in these networks (by replacing ReLU) to improve representation capability with low computational cost. 

\section{Dynamic ReLU}
Dynamic ReLU (DY-ReLU) is a \textbf{dynamic} piecewise function, of which parameters are input dependent. It does NOT increase either the depth or the width of the network, but increases the model capability efficiently with negligible extra computational cost.
This section is organized as follows. We firstly introduce the generic dynamic activation. Then, we present the mathematical definition of DY-ReLU, and how to implement it. Finally, we compare it with prior work.

\subsection{Dynamic Activation}
For a given input vector (or tensor) $\bm{x}$, the dynamic activation is defined as a function $f_{\bm{\theta}(\bm{x})}(\bm{x})$ with learnable parameters $\bm{\theta}(\bm{x})$, which \textit{adapt to the input $\bm{x}$}.
As shown in Figure \ref{fig:overview}, it includes two functions: 
\begin{enumerate}
    \itemsep0.3em
    \item \textit{hyper function} $\bm{\theta}(\bm{x})$: that computes parameters for the activation function.
    \item \textit{activation function} $f_{\bm{\theta}(\bm{x})}(\bm{x})$: that uses the parameters $\bm{\theta}(\bm{x})$ to generate activation for all channels.
\end{enumerate}
Note that the hyper function encodes the global context of all input elements ($x_c \in \bm{x}$) to determine the appropriate activation function. This enables significantly more representation power than its static counterpart,
especially for light-weight models (e.g. MobileNet).  Next, we will discuss dynamic ReLU.

\subsection{Definition and Implementation of Dynamic ReLU}\label{sec:dy-relu}
\noindent \textbf{Definition:} Let us denote the traditional or static ReLU as $\bm{y}=\max\{\bm{x}, 0\}$, where $\bm{x}$ is the input vector. For the input $x_c$ at the $c^{th}$ channel, the activation is computed as $y_c=\max\{x_c, 0\}$. 
ReLU can be generalized to a parametric piecewise linear function $y_c=\max_k\{a_c^kx_c+b_c^k\}$. We propose dynamic ReLU to further extend this piecewise linear function from static to dynamic by adapting $a_c^k$, $b_c^k$ based upon all input elements $\bm{x}=\{x_c\}$ as follows:
\begin{equation}
y_c = f_{\bm{\theta}(\bm{x})}(x_c) = \max_{1 \leq k \leq K}\{a_c^k(\bm{x})x_c+b_c^k(\bm{x})\},
\label{eq:dyrelu}
\end{equation}
where the coefficients ($a_c^k$, $b_c^k$) are the output of a hyper function $\bm{\theta}(\bm{x})$ as:
\begin{equation}
     [a_1^1,\dots,a_C^1, \dots, a_1^K,\dots,a_C^K, b_1^1,\dots,b_C^1,\dots, b_1^K,\dots,b_C^K]^T = \bm{\theta}(\bm{x}),
\label{eq:dyrelu-coeff}
\end{equation}
where $K$ is the number of functions, and $C$ is the number of channels. Note that the activation parameters ($a_c^k$, $b_c^k$) are not only related to its corresponding input $x_c$, but also related to other input elements $x_{j \neq c}$. 

\noindent \textbf{Implementation of hyper function $\bm{\theta}(\bm{x})$}: 
We use a light-weight network to model the hyper function that is similar to Squeeze-and-Excitation (SE) \cite{Hu_2018_CVPR}. For an input tensor $\bm{x}$ with dimension $C\times H \times W$, 
the spatial information is firstly squeezed by global average pooling. It is then followed by two fully connected layers (with a ReLU between them) and a normalization layer. 
The output has $2KC$ elements, corresponding to the \textbf{residual} of $a_{1:C}^{1:K}$ and $b_{1:C}^{1:K}$, which are denoted as $\Delta a_{1:C}^{1:K}$ and $\Delta b_{1:C}^{1:K}$. We simply use $2\sigma (x)-1$ to normalize the residual between -1 to 1, where $\sigma(x)$ denotes sigmoid function. The final output is computed as the sum of initialization and residual as follows: 
\begin{equation}
    a_c^k(\bm{x})=\alpha^k+\lambda_a\Delta a_c^k(\bm{x}), \: b_c^k(\bm{x})=\beta^k+\lambda_b\Delta b_c^k(\bm{x}),
\label{eq:dyrelu-hyper-b}
\end{equation}
where $\alpha^k$ and $\beta^k$ are initialization values of $a_c^k$ and $b_c^k$, respectively. $\lambda_a$ and $\lambda_b$ are scalars that control the range of residual. $\alpha^k$, $\beta^k$, $\lambda_a$ and $\lambda_b$ are hyper parameters. For the case of $K=2$, the default values are $\alpha^1=1$, $\alpha^2=\beta^1=\beta^2=0$, corresponding to static ReLU. The default $\lambda_a$ and $\lambda_b$ are 1.0 and 0.5, respectively.

\subsection{Relation to Prior Work}
\begin{table}[t]
	\begin{center}
		\footnotesize
		\begin{tabular}{l|c|c|c|l}
			\specialrule{.1em}{.05em}{.05em} 
			 &  &	Type & $K$ &  relation to DY-ReLU \\
			\specialrule{.1em}{.05em}{.05em} 
				\begin{tabular} {@{}c@{}} \\ ReLU \cite{NairH10Relu,JarrettKRL09Relu} \\ \end{tabular} &
			    \begin{minipage}{0.1\linewidth}
			    	\includegraphics[width=1.0\linewidth]{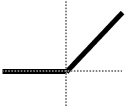}
			    \end{minipage} &
			    \begin{tabular} {@{}c@{}} \\ static \\ \end{tabular} &
			    \begin{tabular} {@{}c@{}} \\ 2 \\ \end{tabular} &
			    \begin{tabular} {@{}l@{}} special case \\ $a_c^1(x)=1,\: b_c^1(x)=0$ \\  $a_c^2(x)=0,\: b_c^2(x)=0$ \\ \end{tabular} \\
			\hline
				\begin{tabular} {@{}c@{}} \\ LeakyReLU \cite{Maas13rectifiernonlinearities} \\ \end{tabular} &
				\begin{minipage}{0.1\linewidth}
					\includegraphics[width=1.0\linewidth]{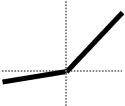}
				\end{minipage} &
				\begin{tabular} {@{}c@{}} \\ static \\ \end{tabular} &
				\begin{tabular} {@{}c@{}} \\ 2 \\ \end{tabular} &
				\begin{tabular} {@{}l@{}} special case \\ $a_c^1(x)=1,\: b_c^1(x)=0$ \\  $a_c^2(x)=\alpha,\: b_c^2(x)=0$ \\ \end{tabular} \\
			\hline
				\begin{tabular} {@{}c@{}} \\ PReLU \cite{he_2015_prelu} \\ \end{tabular} &
				\begin{minipage}{0.11\linewidth}
					\includegraphics[width=1.0\linewidth]{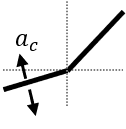}
				\end{minipage} &
				\begin{tabular} {@{}c@{}} \\ static \\ \end{tabular} &
				\begin{tabular} {@{}c@{}} \\ 2 \\ \end{tabular} &
				\begin{tabular} {@{}l@{}} special case \\ $a_c^1(x)=1,\:\: b_c^1(x)=0$ \\  $a_c^2(x)=a_c,\: b_c^2(x)=0$ \\ \end{tabular} \\
			\hline
				\begin{tabular} {@{}c@{}} \\ SE \cite{Hu_2018_CVPR} \\ \end{tabular} &
				\begin{minipage}{0.16\linewidth}
					\includegraphics[width=1.0\linewidth]{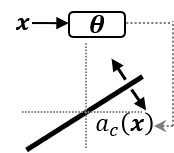}
				\end{minipage} &
				\begin{tabular} {@{}c@{}} \\ dynamic \\ \end{tabular} &
				\begin{tabular} {@{}c@{}} \\ 1 \\ \end{tabular} &
				\begin{tabular} {@{}l@{}} special case \\$a_c^1(x)=a_c(x),\: b_c^1(x)=0$ \\ $0 \leq a_c(x) \leq 1$ \\ \end{tabular} \\
			\hline
				\begin{tabular} {@{}c@{}} \\ Maxout \cite{goodfellow2013maxout} \\ \end{tabular} &
				\begin{minipage}{0.14\linewidth}
					\includegraphics[width=1.0\linewidth]{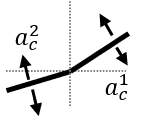}
				\end{minipage} &
				\begin{tabular} {@{}c@{}} \\ static \\ \end{tabular} &
				\begin{tabular} {@{}c@{}} \\ 1,2,3,... \\ \end{tabular} &
				\begin{tabular} {@{}l@{}} DY-ReLU is a dynamic \\ and efficient Maxout. \\ \end{tabular} \\
			\specialrule{.1em}{.05em}{.05em}
				\begin{tabular} {@{}c@{}} \\ DY-ReLU \\ \end{tabular} &
				\begin{minipage}{0.18\linewidth}
					\includegraphics[width=1.0\linewidth]{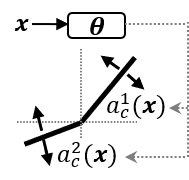}
				\end{minipage} &
				\begin{tabular} {@{}c@{}} \\ dynamic \\ \end{tabular} &
				\begin{tabular} {@{}c@{}} \\ 1,2,3,... \\ \end{tabular} &
				\begin{tabular} {@{}l@{}} \\ identical \\ \end{tabular} \\
			\specialrule{.1em}{.05em}{.05em} 
		\end{tabular}
	\end{center}
	\caption{Relation to prior work. ReLU, LeakyReLU, PReLU and SE are special cases of DY-ReLU. DY-ReLU is a \textit{dynamic} and \textit{efficient} version of Maxout. $\alpha$ in LeakyReLU is a small number (e.g. 0.01). $a_c$ in PReLU is a parameter to learn.}
	\label{table:relation-prior}
\end{table}

Table \ref{table:relation-prior} shows the relationship between DY-ReLU and prior work. The three special cases of DY-ReLU are equivalent to ReLU \cite{NairH10Relu,JarrettKRL09Relu}, LeakyReLU \cite{Maas13rectifiernonlinearities} and PReLU \cite{he_2015_prelu}, where the hyper function becomes static. 
%
SE \cite{Hu_2018_CVPR} is another special case of DY-ReLU, with a \textit{single} linear function $K=1$ and zero intercept $b_c^1=0$.

DY-ReLU is a \textit{dynamic} and  \textit{efficient} Maxout \cite{goodfellow2013maxout}, with significantly less computations but even better performance.
Different with Maxout that requires multiple ($K$) convolutional kernels, DY-ReLU applies $K$ \textit{dynamic} linear transforms on the results of a \textit{single} convolutional kernel, and outputs the maximum of them. This results in much less computations and even better performance.

\section{Variations of Dynamic ReLU}
\begin{figure}[t!]
	\begin{center}
		\includegraphics[width=1.0\linewidth]{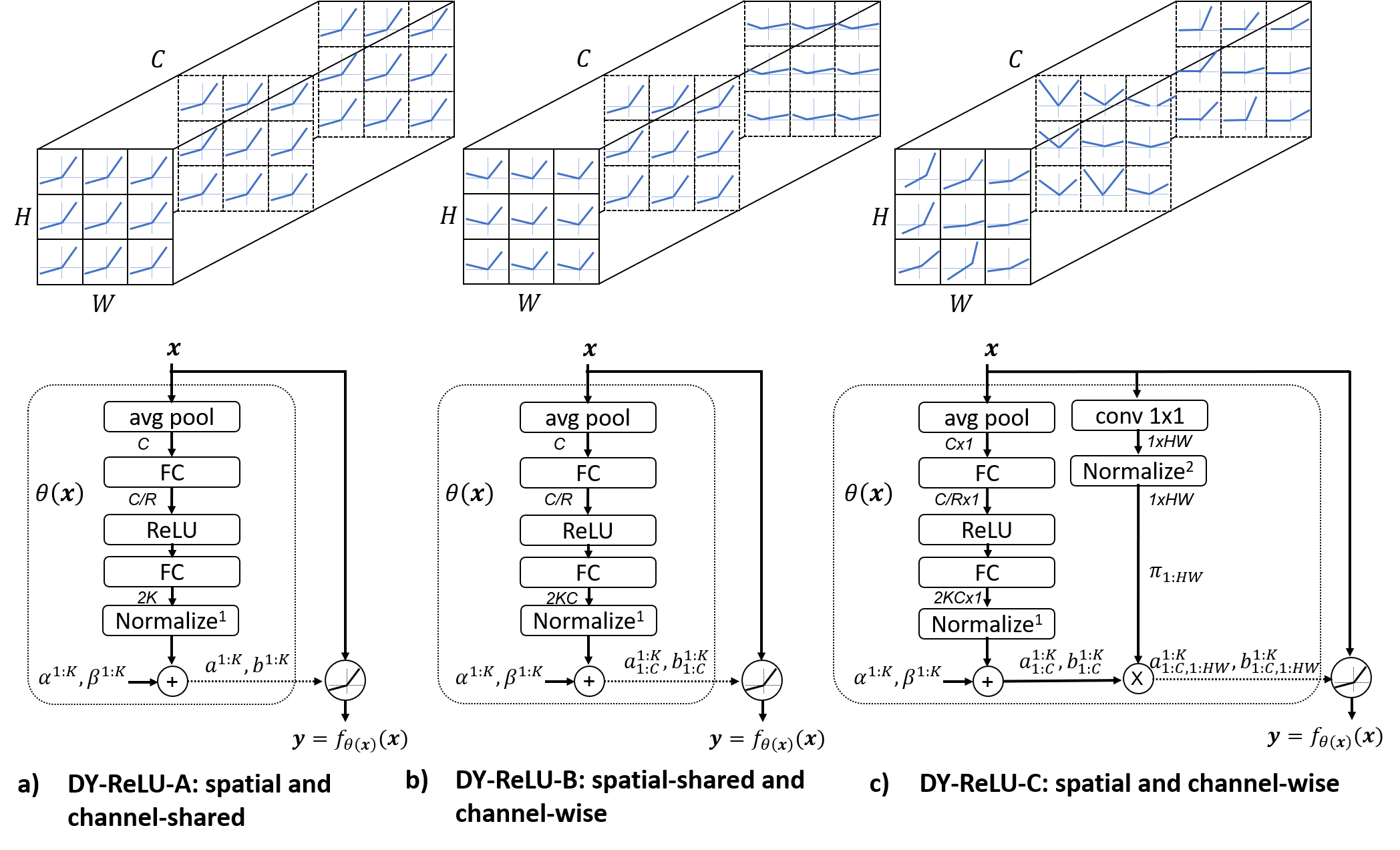}
	\end{center}
	\caption{Three DY-ReLU variations. They have different ways of sharing activation functions. The top row illustrates the piecewise linear function across spatial locations and channels, and the bottom row shows the network structure for the hyper function. Note that the first FC layer reduces the dimension by $R$, which is a hyper parameter.}
	\label{fig:sharing}
\end{figure}

In this section, we introduce another two variations of dynamic ReLU in addition to the option discussed in section \ref{sec:dy-relu}. These three options have different ways of sharing activation functions as follows:
\begin{enumerate}
    \itemsep0.3em
    \item[] \textbf{DY-ReLU-A}: the activation function is \textit{spatial and channel-shared}.
    \item[] \textbf{DY-ReLU-B}: the activation function is \textit{spatial-shared and channel-wise}.
    \item[] \textbf{DY-ReLU-C}: the activation function is \textit{spatial and channel-wise}.
\end{enumerate}
DY-ReLU-B has been discussed in section \ref{sec:dy-relu}.
%
\subsection{Network Structure and Complexity}
The network structures of three variations are shown in Fig. \ref{fig:sharing}. The detailed explanation is discussed as follows:

\noindent \textbf{DY-ReLU-A (Spatial and Channel-shared)}: the same piecewise linear activation function is shared across all spatial positions and channels. Its hyper function has similar network structure (shown in Fig. \ref{fig:sharing}-(a)) to DY-ReLU-B, except the number of outputs is reduced to $2K$. Compared to DY-ReLU-B, DY-ReLU-A has less computational cost, but less representation capability. 

\noindent \textbf{DY-ReLU-B (Spatial-shared and Channel-wise)}: the implementation details are introduced in section \ref{sec:dy-relu} and the network structure is shown in Fig. \ref{fig:sharing}-(b). The hyper function outputs $2KC$ parameters ($2K$ per channel).

\noindent \textbf{DY-ReLU-C (Spatial and Channel-wise)}: as shown in Fig. \ref{fig:sharing}-(c), each input element $x_{c,h,w}$ has a unique activation function $\max_k\{a_{c,h,w}^kx_{c,h,w}+b_{c,h,w}^k\}$, where the subscript $_{c,h,w}$ indicates the $c^{th}$ channel at the $h^{th}$ row and $w^{th}$ column of the feature map that has dimension $C \times H \times W$. This introduces an issue that the output dimension is too large ($2KCHW$), resulting in significantly more parameters in the fully connected layer. We address it by decoupling spatial locations from channels. Specifically, another branch for computing spatial attention $\pi_{h,w}$ is introduced.
The final output is computed as the product of channel-wise parameters ($[a_{1:C}^{1:K}, b_{1:C}^{1:K}]^T$) and spatial attentions ($[\pi_{1:HW}]$). The spatial attention branch is simple, including a $1\times 1$ convolution with a single output channel and a normalization that is a softmax function with upper cutoff as follows:
\begin{align}
\pi_{h, w}=\min\{\frac{\gamma\exp(z_{h,w}/\tau)}{\sum_{h, w}\exp(z_{h,w}/\tau)}, 1\},
\label{eq:gumbel}
\end{align}
where $z_{h,w}$ is the output of $1\times 1$ convolution, $\tau$ is the temperature, and $\gamma$ is a scalar. The softmax is scaled up by $\gamma$ is to prevent gradient vanishing. We empirically set $\gamma=\frac{HW}{3}$, making the average attention $\pi_{h, w}$ to $\frac{1}{3}$. A large temperature ($\tau=10$) is used to prevent sparsity during the early training stage. The upper bound $1$ constrains the attention between zero and one. 

\noindent \textbf{Computational Complexity}: DY-ReLU is computationally efficient. It includes four components: (a) average pooling, (b) the first FC layer (with ReLU), (c) the second FC layer (with normalization), and (d) piecewise function $f_{\bm{\theta}(\bm{x})}(\bm{x})$. 
For a feature map with dimension $C\times H \times W$, all three DY-ReLU variations share complexity for average pooling $O(CHW)$, the first FC layer $O(C^2/R)$ and piecewise function $O(CHW)$. The second FC layer has complexity $O(2KC/R)$ for DY-ReLU-A and $O(2KC^2/R)$ for DY-ReLU-B and DY-ReLU-C. Note that DY-ReLU-C spends additional $O(CHW)$ on computing spatial attentions.
In most of the layers of MobileNet and ResNet, DY-ReLU has much less computation than a $1\times1$ convolution, which has complexity $O(C^2HW)$.


\subsection{Ablations} \label{sec:dy-relu-var-ablation}
Next, we study the three DY-ReLU variations on image classification and keypoint detection. Our goal is to understand their differences when performing different tasks. The details of datasets, implementation and training setup will be shown later in the next section.

\begin{table}[t]
\parbox{.40\linewidth}{
\centering
\small
\begin{tabular}{l|l}
        \specialrule{.1em}{.05em}{.05em}
         & Top-1 \\
        \specialrule{.1em}{.05em}{.05em}
        ReLU & $60.32 \pm 0.13$ \\
        \cline{1-2}
        DY-ReLU-A & $63.28 \pm 0.12_{(2.96)}$ \\
        DY-ReLU-B & $\bm{66.36 \pm 0.12}_{(\bm{6.04})}$ \\
        DY-ReLU-C & $66.31 \pm 0.14_{(5.99)}$ \\
        \specialrule{.1em}{.05em}{.05em}
        \multicolumn{2}{c}{}\\
    \end{tabular}
    \caption{Comparing three DY-ReLU variations on Imagenet \cite{deng2009imagenet} classification. MobileNetV2 with width multiplier $\times 0.35$ is used. The mean and standard deviations of three runs are shown. The numbers in brackets denote the performance improvement over the baseline. Channel-wise variations (DY-ReLU-B and DY-ReLU-C) are more effective than the channel-shared (DY-ReLU-A). Spatial-wise (DY-ReLU-C) does NOT introduce additional improvement.}
    \label{table:cls-ablation-ABC}
    }
\hfill
\parbox{.56\linewidth}{
\centering
\small
\begin{tabular}{l|l|l } 
			\specialrule{.1em}{.05em}{.05em} 
			Backbone & Head & AP \\
			\specialrule{.1em}{.05em}{.05em}
			ReLU      & ReLU    & $59.26\pm0.21$  \\
			\hline
			DY-ReLU-A      & ReLU   & $58.97\pm0.15_{(-0.29)}$ \\
			DY-ReLU-B     & ReLU    & $61.76\pm0.27_{(+2.50)}$ \\
			DY-ReLU-C      & ReLU   & $62.23\pm0.32_{(+2.97)}$ \\
			\hline
			ReLU      & DY-ReLU-A   & $57.12\pm0.25_{(-2.14)}$ \\
			ReLU      & DY-ReLU-B   & $58.72\pm0.35_{(-0.54)}$ \\
			ReLU      & DY-ReLU-C   & $61.03\pm0.11_{(+1.77)}$ \\
			
			\hline
			DY-ReLU-C     & DY-ReLU-C & $\bm{63.27\pm0.15}_{(\bm{+4.01})}$ \\
			\specialrule{.1em}{.05em}{.05em} 
			 \multicolumn{3}{c}{}\\
		\end{tabular}
    \caption{Comparing three DY-ReLU variations on COCO \cite{lin2014microsoft} keypoint detection. We use MobileNetV2 $\times0.5$ as backbone and use up-sampling and inverted residual bottleneck blocks \cite{Chen2019DynamicCA} in the head. The mean and standard deviations of three runs are shown. The numbers in brackets denote the performance improvement over the baseline. Channel-wise variations (DY-ReLU-B and DY-ReLU-C) are more effective in the backbone and the spatial-wise variation (DY-ReLU-C) is more effective in the head.}
    \label{table:pose-ablation}
}
\end{table}
The comparison among three DY-ReLU variations on ImageNet \cite{lin2014microsoft} classification is shown in Table \ref{table:cls-ablation-ABC}. MobileNetV2 $\times 0.35$ is used. Although all three variations achieve improvement from the baseline, \textbf{\textit{channel-wise DY-ReLUs (variation B and C) are clearly better than the channel-shared DY-ReLU (variation A)}}. Variation B and C have similar accuracy, showing that spatial-wise is not critical for image classification. 

Table \ref{table:pose-ablation} shows the comparison on COCO keypoint detection. Similar to image classification, \textbf{\textit{channel-wise variations (B and C) are better than channel-shared variation A in the backbone}}. In contrast, \,\textbf{\textit{the spatial-wise variation C is critical in the head}}. Using DY-ReLU-C in both backbone and head achieves 4 AP improvement. We also observe that the performance is even worse than the baseline if we use DY-ReLU-A in the backbone or use DY-ReLU-A and DY-ReLU-B in the head.
We believe the spatially-shared hyper function in DY-ReLU-A or DY-ReLU-B is difficult to learn when dealing with spatially sensitive task (e.g. distinguishes body joints in pixel level), especially in the head that has higher resolutions. This difficulty can be effectively alleviated by making hyper function spatial-wise, which encourages learning different activation functions at different positions. We observe that the training converges much faster when using spatial attention in the head network.

%
%
%



Base upon these ablations, we use DY-ReLU-B for ImageNet classification and use DY-ReLU-C for COCO keypoint detection in the next section.

\section{Experimental Results}
In this section, we present experimental results on image classification and single person pose estimation to demonstrate the effectiveness of DY-ReLU. We also report ablation studies to analyze different components of our approach.

\subsection{ImageNet Classification}
We use ImageNet \cite{deng2009imagenet} for all classification experiments. ImageNet has 1000 classes, including 1,281,167 images for training and 50,000 images for validation. We evaluate DY-ReLU on three CNN architectures (MobileNetV2 \cite{sandler2018mobilenetv2}, MobileNetV3 \cite{howard2019mbnetv3} and ResNet \cite{he2016deep}). We replace their default activation functions (ReLU in ResNet and MobileNetV2, ReLU/hswish/SE in MobileNetV3) with DY-ReLU. The main results are obtained by using spatial-shared and channel-wise DY-ReLU-B with two piecewise linear functions ($K=2$). Note that MobileNetV2 and V3 have no activation after the last convolution layer in each block, where we add DY-ReLU with $K=1$. 
The batch size is 256. We use different training setups for the three architectures as follows:

\noindent \textbf{Training setup for MobileNetV2:} The initial learning rate is 0.05, and is scheduled to arrive at zero within a single cosine cycle. All models are trained using SGD optimizer with 0.9 momentum for 300 epochs. The label smoothing 0.1 is used. 
We use weight decay 2e-5 and dropout 0.1 for width $\times 0.35$, and increase weight decay 3e-5 and dropout 0.2 for width $\times 0.5$, $\times 0.75$, $\times 1.0$. Random cropping/flipping and color jittering are used for all width multipliers. Mixup \cite{zhang2018mixup} is used for width $\times 1.0$ to prevent overfitting.

\noindent \textbf{Training setup for MobileNetV3:} The initial learning rate is 0.1 and is scheduled to arrive at zero within a single cosine cycle. The weight decay is 3e-5 and label smoothing is 0.1. We use SGD optimizer with 0.9 momentum for 300 epochs. 
We use dropout rate of 0.1 and 0.2 before the last layer for MobileNetV3-Small and MobileNetV3-Large respectively. We use more data augmentation (color jittering and Mixup \cite{zhang2018mixup}) for MobileNetV3-Large.

\noindent \textbf{Training setup for ResNet:} The initial learning rate is 0.1 and drops by 10 at epoch 30, 60. The weight decay is 1e-4. All models are trained using SGD optimizer with 0.9 momentum for 90 epochs. We use dropout rate 0.1 before the last layer and label smoothing for ResNet-18, ResNet-34 and ResNet-50.


\begin{table}[t]
	\begin{center}
	    \footnotesize
		\begin{tabular}{r|r|r| r|l l}
			\specialrule{.1em}{.05em}{.05em} 
			Network & Activation& \#Param & MAdds &Top-1 &	Top-5   \\
			
			\hline
			MobileNetV2 $\times1.0$  & ReLU & 3.5M & 300.0M & 72.0  & 91.0  \\
			                         & DY-ReLU & 7.5M & 315.5M & 76.2$_{(4.2)}$ & 93.1$_{(2.1)}$	  \\
			\hline
			MobileNetV2 $\times0.75$  & ReLU & 2.6M & 209.0M & 69.8 & 89.6 		 \\
			                          & DY-ReLU & 5.0M & 221.7M  & 74.3$_{(4.5)}$ & 91.7$_{(2.1)}$		 \\
			\hline
			MobileNetV2 $\times0.5$  & ReLU & 2.0M & 97.0M & 65.4 & 86.4 		 \\
			                         & DY-ReLU & 3.1M & 104.5M & 70.3$_{(4.9)}$ & 89.3$_{(2.9)}$		 \\
			\hline
			MobileNetV2 $\times0.35$ & ReLU & 1.7M & 59.2M  & {60.3}& {82.9} 		 \\
			                         & DY-ReLU & 2.7M & 65.0M & 66.4$_{(6.1)}$ & 86.5$_{(3.6)}$ 		 \\
			
			\specialrule{.1em}{.05em}{.05em} 
			
			MobileNetV3-Large  & ReLU/SE/HS & 5.4M & 219.0M & 75.2 & 92.2   \\
			 & DY-ReLU & 9.8M & 230.5M & 75.9$_{(0.7)}$ & 92.7$_{(0.5)}$	 		 \\

		    \hline
			MobileNetV3-Small  & ReLU/SE/HS & 2.9M & 66.0M & 67.4 & 86.4   \\
			 & DY-ReLU & 4.0M & 68.7M & 69.7$_{(2.3)}$ & 88.3$_{(1.9)}$	 		 \\

            \specialrule{.1em}{.05em}{.05em} 
			ResNet-50 & ReLU & 23.5M & 3.86G & 76.2  & 92.9  \\
			& DY-ReLU& 27.6M & 3.88G & 77.2$_{(1.0)}$ & 93.4$_{(0.5)}$	  \\
			\hline
			ResNet-34 & ReLU & 21.3M & 3.64G & 73.3  & 91.4  \\
			& DY-ReLU& 24.5M & 3.65G & 74.4$_{(1.1)}$ & 92.0$_{(0.6)}$	  \\
			
			\hline
			ResNet-18 & ReLU & 11.1M & 1.81G & 69.8  & 89.1  \\
			& DY-ReLU& 12.8M & 1.82G & 71.8$_{(2.0)}$ & 90.6$_{(1.5)}$	  \\
			\hline
			ResNet-10 & ReLU & 5.2M & 0.89G & 63.0  & 84.7  \\
			& DY-ReLU & 6.3M & 0.90G & 66.3$_{(3.3)}$ & 86.7$_{(2.0)}$	  \\
			\specialrule{.1em}{.05em}{.05em} 
		\end{tabular}
	\end{center}
	\caption{Comparing DY-ReLU with baseline activation functions (ReLU, SE or h-swish, denoted as HS) on ImageNet \cite{deng2009imagenet} classification in three network architectures. DY-ReLU-B with $K=2$ linear functions is used. Note that SE blocks are removed when using DY-ReLU in MobileNetV3. The numbers in brackets denote the performance improvement over the baseline. DY-ReLU outperforms its counterpart for all networks.}
	\label{table:imagenet-cls-result}
\end{table}

\begin{table}[t]
	\begin{center}
	    \footnotesize
		\begin{tabular}{l|c|r| r|l| r| r|l}
			\specialrule{.1em}{.05em}{.05em} 
			& & \multicolumn{3}{c|}{MobileNetV2 $\times$0.35} & \multicolumn{3}{c}{MobileNetV2 $\times$1.0} \\
			\cline{3-8}
			Activation & K & \#Param & MAdds &Top-1 & \#Param & MAdds &Top-1  \\
			\specialrule{.1em}{.05em}{.05em}
			ReLU & 2 & 1.7M & 59.2M  & {60.3} 	& 3.5M & 300.0M  & {72.0}	 \\
			\hline
			
			RReLU \cite{Xu15empiricalactivaiton} & 2 & 1.7M & 59.2M  & {60.0}$_{(-0.3)}$& 3.5M & 300.0M  & {72.5}$_{(+0.5)}$ 		 \\
			LeakyReLU \cite{Maas13rectifiernonlinearities} & 2 & 1.7M & 59.2M  & {60.9}$_{(+0.6)}$& 3.5M & 300.0M  & {72.7}$_{(+0.7)}$ 		 \\
			PReLU \cite{he_2015_prelu} & 2 & 1.7M & 59.2M  & {63.1}$_{(+2.8)}$& 3.5M & 300.0M  & {73.3}$_{(+1.3)}$ 		 \\
			SE\cite{Hu_2018_CVPR}+ReLU & 2 & 2.1M & 62.0M  & {62.8}$_{(+2.5)}$& 5.1M & 307.5M  & {74.2}$_{(+2.2)}$ 		 \\
			Maxout \cite{goodfellow2013maxout} & 2 & 2.1M & 106.6M & 64.9$_{(+4.6)}$ & 5.7M & 575.8M & 75.1$_{(+3.1)}$		 \\
			Maxout \cite{goodfellow2013maxout}& 3 & 2.4M & 157.6M & 65.4$_{(+5.1)}$ & 7.8M & 860.2M & 75.8$_{(+3.8)}$		 \\
			\hline
			DY-ReLU-B & 2 & 2.7M & 65.0M & 66.4$_{(+6.1)}$ & 7.5M & 315.5M & \textbf{76.2}$_{(\bm{+4.2})}$		 \\
			DY-ReLU-B & 3 & 3.1M & 67.8M & \textbf{66.6}$_{(\bm{+6.3})}$ & 9.2M & 322.8M & \textbf{76.2}$_{(\bm{+4.2})}$ 		 \\
			
			\specialrule{.1em}{.05em}{.05em} 
		\end{tabular}
	\end{center}
	\caption{Comparing DY-ReLU with related activation functions on ImageNet \cite{deng2009imagenet} classification. MobileNetV2 with width multiplier $\times 0.35$ and $\times 1.0$ are used. We use spatial-shared and channel-wise DY-ReLU-B with $K=2,3$ linear functions. The numbers in brackets denote the performance improvement over the baseline. DY-ReLU outperforms all prior work including Maxout, which has significantly more computations.}
	\label{table:imagenet-cls-prior-work}
\end{table}

\noindent \textbf{Main Results:} We compare DY-ReLU with its static counterpart in three CNN architectures (MobileNetV2, MobileNetV3 and ResNet) in Table \ref{table:imagenet-cls-result}. 
Without bells and whistles, DY-ReLU outperforms its static counterpart by a clear margin for all three architectures, with small extra computational cost ($\sim5\%$). DY-ReLU gains more than 1.0\% top-1 accuracy in ResNet and gains more than 4.2\% top-1 accuracy in MobileNetV2. For the state-of-the-art MobileNetV3, our DY-ReLU outperforms the combination of SE and h-swish (key contributions of MobileNetV3). The top-1 accuracy is improved by 2.3\% and 0.7\% for MobileNetV3-Small and MobileNetV3-Large, respectively. 
Note that DY-ReLU achieves more improvement for smaller models (e.g. MobileNetV2 $\times 0.35$, MobileNetV3-Small, ResNet-10). This is because the smaller models are underfitted due to their model size, and dynamic ReLU significantly boosts their representation capability.

The comparison between DY-ReLU and prior work is shown in Table \ref{table:imagenet-cls-prior-work}. Here we use MobileNetV2 ($\times 0.35$ and $\times 1.0$), and replace ReLU with different activation functions in prior work. Our method outperforms all prior work with a clear margin, including Maxout that has significantly more computational cost. This demonstrates that DY-ReLU not only has more representation capability, but also is computationally efficient. 


\subsection{Inspecting DY-ReLU: Is It Dynamic?}
\begin{figure}[t!]
	\begin{center}
		\includegraphics[width=1.0\linewidth]{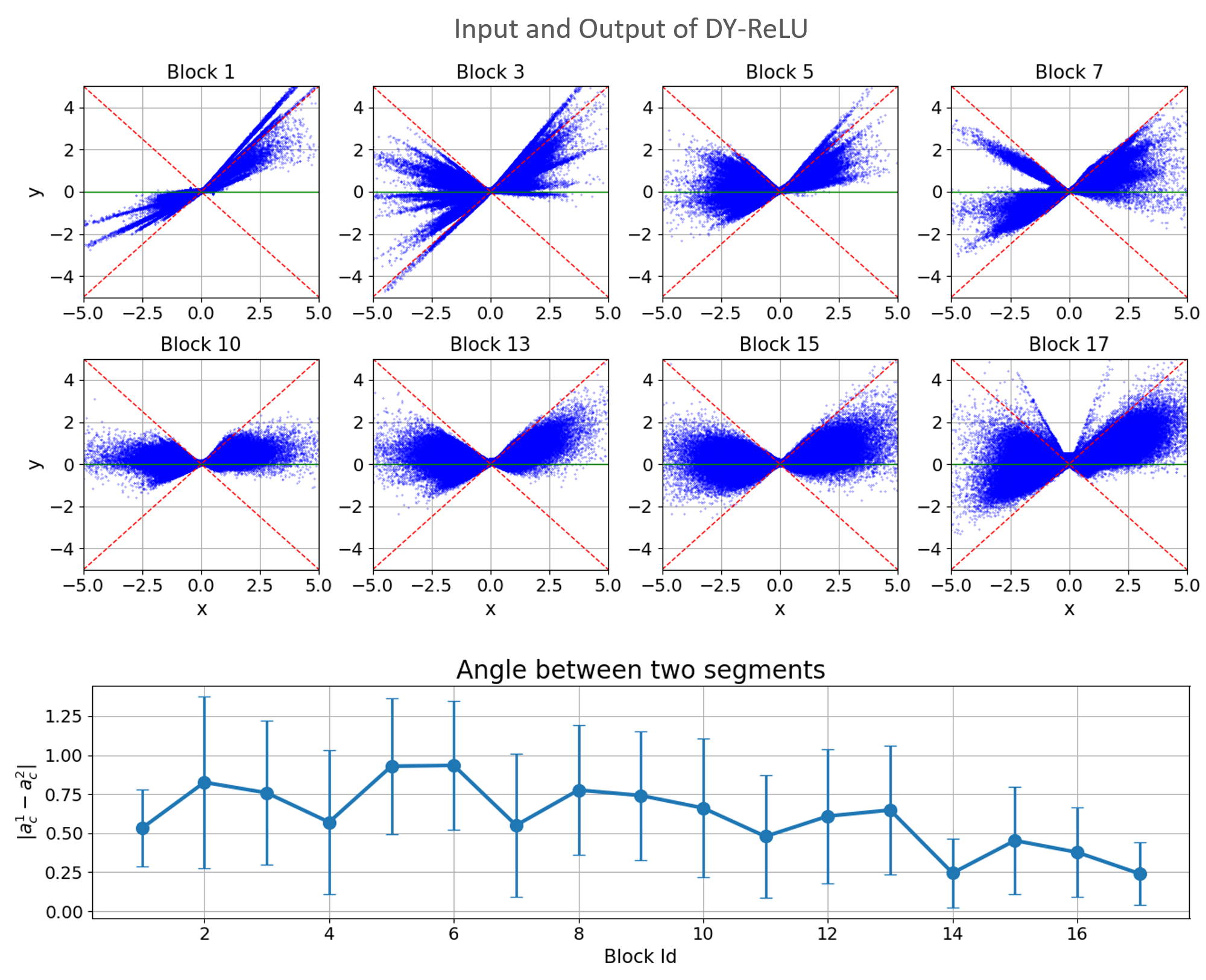}
	\end{center}
	\caption{\textbf{Top:} plots of input and output values of DY-ReLU in a well trained model (using MobileNetV2 $\times 0.35$) over 50,000 validation images in ImageNet \cite{deng2009imagenet}. We choose the dynamic ReLU after the depthwise convolution in every other mobile block. Block 1 corresponds to the lowest block, and Block 17 corresponds to the highest block. The two \textcolor{red}{red lines} correspond to $y=x$ and $y=-x$, respectively. \textbf{Bottom:} Angle (or slope difference $|a_c^1-a_c^2|$) between two segments in DY-ReLU across blocks. The bending of the activation functions decreases from low levels to high levels.  Best viewed in color.}
	\label{fig:dy-relu-vis}
\end{figure}
We check if DY-ReLU is dynamic by examining its input and output over multiple images. Different activation values ($y$) across different images for a given input value (e.g. $x=0.5$) is expected to differentiate from static ReLU, which has a fixed output (e.g. $y=0.5$ when $x=0.5$). 

Fig. \ref{fig:dy-relu-vis}-(Top) plots the input and output values of DY-ReLU at different blocks (from low level to high level) for 50,000 validation images in ImageNet \cite{deng2009imagenet}. Clearly, the learnt DY-ReLU is dynamic over features as activation values ($y$) vary in a range (that blue dots cover) for a given input $x$. The dynamic range varies across different blocks, indicating different dynamic functions learnt across levels. We also observe many positive activations for negative inputs. Statistically, 51\% of DY-ReLU have segments with either negative slope or slope above 1, and 37\% of DY-ReLU have at least one segment with intercept more than 0.05. These cases cannot be handled by ReLU, SE or MaxOut of two SEs.

We also analyzed the angle between two segments in DY-ReLU (i.e. slope difference 
$|a_c^1-a_c^2|$). The slope difference decreases from lower to higher levels (shown in Fig. \ref{fig:dy-relu-vis}-(Bottom)). This indicates that the activation functions tend to have lower bending in higher levels.

\subsection{Ablation Studies on ImageNet}
We run a number of ablations to analyze DY-ReLU. We focus on spatial-shared and channel-wise DY-ReLU-B, and use MobileNetV2 $\times 0.35$ for all ablations. By default, the number of linear functions in DY-ReLU is set as $K=2$. The initialization values of slope and intercept are set as $\alpha^1=1$, $\alpha^2=\beta^1=\beta^2=0$. The range of slope and intercept are set as $\lambda_a=1$ and $\lambda_b=0.5$, respectively. The reduction ratio of the first FC layer in the hyper function is set as $R=8$. 

\begin{table}[t]
    \begin{center}
    \small
    \begin{tabular}{l|c|c|l|cc}
        \specialrule{.1em}{.05em}{.05em}
        & $K$ & intercept $b_c^k$ & Activation Function &  Top-1 & Top-5 \\
        \specialrule{.1em}{.05em}{.05em}
        ReLU & 2 & & $\max\{x_c, 0\}$ & 60.3 & 82.9 \\
        \hline
        & 2 & & $\max\{a_c(\bm{x})x_c, 0\}$ &  63.8 & 85.1 \\
        & 2 & \checkmark & $\max\{a_c(\bm{x})x_c+b_c(\bm{x}), 0\}$ & 64.0 & 85.2  \\
        DY-ReLU & 2&&$\max_{k=1}^2\{a_c^k(\bm{x})x_c\}$  &65.7 & 86.2 \\
        & 2 & \checkmark & $\max_{k=1}^2(a_c^k(\bm{x})x_c+b_c^k(\bm{x})\}$ &66.4 & 86.5 \\
        & 3&&$\max_{k=1}^3\{a_c^k(\bm{x})x_c\}$ & 65.9 & 86.3 \\
        & 3 & \checkmark & $\max_{k=1}^3\{a_c^k(\bm{x})x_c+b_c^k(\bm{x})\}$ &66.6 & 86.8 \\
        \specialrule{.1em}{.05em}{.05em}
    \end{tabular}
    \end{center}
    \caption{\textbf{Different dynamic piecewise functions} evaluated on ImageNet classification. MobileNetV2 $\times 0.35$ is used.}
    \label{table:ablation-piecewise-linear}
\end{table}

\noindent \textbf{Dynamic Piecewise Functions:} Table \ref{table:ablation-piecewise-linear} shows the classification accuracy using different piecewise functions. The major gain is due to making ReLU dynamic. Specifically, making the first segment dynamic boosts top-1 accuracy from 60.3\% to 63.8\%. Making the second segment dynamic gains additional 1.9\%. The intercept $b_c^k$ is helpful consistently. The gap between $K=2$ and $3$ is small. In most of DY-ReLU with $K=3$ segments, 2 of the 3 segments have similar slopes.


\begin{table}[t]
\parbox{.46\linewidth}{
\centering
    \small
    \begin{tabular}{c|c@{\hskip 2mm}c@{\hskip 2mm}c|cc}
        \specialrule{.1em}{.05em}{.05em}
         & $A_1$ & $A_2$ & $A_3$ & Top-1 & Top-5 \\
        \specialrule{.1em}{.05em}{.05em}
        ReLU & -- & -- & -- & 60.3 & 82.9\\
        \hline
         & \checkmark & -- & -- & 64.2 & 84.9 \\
         & -- & \checkmark & -- & 65.3 & 85.9 \\
         & -- & -- & \checkmark & 62.7 & 83.8\\
        DY-ReLU & \checkmark & \checkmark & -- & 66.2 & 86.4 \\
         & \checkmark & -- & \checkmark & 64.5 & 85.3 \\
         & -- & \checkmark & \checkmark & 65.9 & 86.2 \\        
         & \checkmark & \checkmark & \checkmark & 66.4 & 86.5 \\
        \specialrule{.1em}{.05em}{.05em}
         \multicolumn{6}{c}{}\\
    \end{tabular}
    \caption{\textbf{DY-ReLU at different layers} evaluated on ImageNet. MobileNetV2 $\times 0.35$ is used. $A_1,A_2,A_3$ indicate activations after three convolution layers in an inverted residual block.}
    \label{table:ablation-diff-layer}
    }
%
\hfill
\parbox{.51\linewidth}{
\centering
    \small
    \begin{tabular}{l|c@{\hskip 2mm}|r|r|cc}
            \specialrule{.1em}{.05em}{.05em}
            & $R$& \#param & MAdds & Top-1 & Top-5 \\
            \specialrule{.1em}{.05em}{.05em}
            ReLU& -- & 1.7M & 59.2M & 60.3 & 82.9 \\
            \hline
            &$64$ & 2.0M & 64.3M & 65.0 & 85.7 \\
            &$32$ & 2.1M & 64.4M & 65.5 & 86.0 \\
            DY-ReLU & $16$ & 2.3M & 64.6M & 65.9 & 86.3 \\
            &$8$ & 2.7M & 65.0M &66.4 & 86.5 \\
            &$4$ & 3.6M & 65.9M & 66.5 & 86.7 \\
            \specialrule{.1em}{.05em}{.05em}
             \multicolumn{6}{c}{}\\
    \end{tabular}
    \caption{\textbf{Different reduction ratios $R$} for the first fully connected layer in the hyper function (see Fig. \ref{fig:sharing}). Evaluation is on ImageNet classification. MobileNetV2 $\times 0.35$ is used. Setting $R=8$ achieves a good trade-off.}
    \label{table:ablation-diff-R}
}
\end{table}

\noindent \textbf{Dynamic ReLU at Different Layers:} Table \ref{table:ablation-diff-layer} shows the classification accuracy for using DY-ReLU at three different layers (after $1\times1$ conv, $3\times3$ depthwise conv, $1\times1$ conv) in an inverted residual block in MobileNetV2 $\times 0.35$. The accuracy is improved if DY-ReLU is used for more layers. Using DY-ReLU for all three layers yields the best accuracy. If only one layer is allowed to use DY-ReLU, using it after $3\times3$ depth-wise convolution yields the best performance. 

\noindent \textbf{Reduction Ratio $R$:} The reduction ratio of the first FC layer in the hyper function $\bm{\theta}(\bm{x})$ controls the representation capacity and computational cost of DY-ReLU. The comparison across different reduction ratios is shown in Table \ref{table:ablation-diff-R}. Setting $R=8$ achieves a good trade-off.

\begin{table}[t]
    \begin{center}
    \footnotesize
    \begin{tabular}{ll@{\hskip 3mm}ll@{\hskip 3mm}l}
    \begin{tabular}{r@{\hskip 2mm}r@{\hskip 2mm}cc}
        \specialrule{.1em}{.05em}{.05em}
         $\alpha^1$& $\alpha^2$ & Top-1 & Top-5 \\
        \specialrule{.1em}{.05em}{.05em}
        1.0 & 0.0 & 66.4 & 86.5 \\
        \cline{1-4}
        1.5 & 0.0 & 65.7 & 86.2 \\
        0.5 & 0.0 & 66.1 & 86.3 \\
        0.0 & 0.0 & \multicolumn{2}{c}{not converge}\\
        \hline
        1.0 & -0.5 & 65.2 & 85.5 \\
        1.0 & 0.5 & 66.4 & 86.2 \\
        1.0 & 1.0 & 66.0 & 86.1 \\
        \specialrule{.1em}{.05em}{.05em} \\
        \multicolumn{4}{l}{(a) \textbf{Initialization of $\alpha^k$}. }
    \end{tabular}
& &
    \begin{tabular}{r@{\hskip 2mm} r@{\hskip 2mm} cc}
    \\
    \\
        \specialrule{.1em}{.05em}{.05em}
         $\beta^1$& $\beta^2$ & Top-1 & Top-5 \\
        \specialrule{.1em}{.05em}{.05em}
        0.0 & 0.0 & 66.4 & 86.5 \\
        \cline{1-4}
        -0.1 & 0.0 & 66.4 & 86.5 \\
        0.1 & 0.0 & 66.2 & 86.4   \\
        \hline
        0.0 & -0.1 & 65.8 & 86.2 \\
        0.0 & 0.1 & 65.3 & 85.8 \\
        \specialrule{.1em}{.05em}{.05em} \\
        \multicolumn{4}{l}{(b) \textbf{Initialization of $\beta^k$}.}
    \end{tabular}
& &
    \begin{tabular}{r@{\hskip 5mm}cc}
    \\
    \\
    \\
        \specialrule{.1em}{.05em}{.05em}
         $\lambda_a$& Top-1 & Top-5 \\
        \specialrule{.1em}{.05em}{.05em}
        0.5 & 65.3 & 86.0 \\
        1.0 & 66.4 & 86.5 \\
        2.0 & 66.3 & 86.5 \\
        3.0 & 65.5 & 86.1 \\
        \specialrule{.1em}{.05em}{.05em} \\
        \multicolumn{3}{l}{(c) \textbf{Range of slope $\lambda_a$}.}
    \end{tabular}
    \end{tabular}
    \end{center}
    \caption{Ablations of three hyper parameters in DY-ReLU on Imagenet classification.}
    \label{table:ablation-init-range}
\end{table}

\noindent \textbf{Initialization of Slope} ($\alpha^k$ in Eq (\ref{eq:dyrelu-hyper-b})): As shown in Table \ref{table:ablation-init-range}-(a), the classification accuracy is not sensitive to the initialization values of slopes if the first slope is not close to zero and the second slope is non-negative.

\noindent \textbf{Initialization of Intercept} ($\beta^k$ in Eq (\ref{eq:dyrelu-hyper-b})): the performance is stable (shown in Table \ref{table:ablation-init-range}-(b)) when both intercepts are close to zero. The second intercept is more sensitive than the first one, as it moves the interception of two lines further away from the origin diagonally.

\noindent \textbf{Range of slope} ($\lambda_a$ in Eq (\ref{eq:dyrelu-hyper-b})): Making slope range either too wide or too narrow is not optimal, as shown in Table \ref{table:ablation-init-range}-(c). A good choice is to keep $\lambda_a$ between 1 and 2.

\subsection{COCO Single-Person Keypoint Detection}

We use COCO 2017 dataset \cite{lin2014microsoft} to evaluate dynamic ReLU on single-person keypoint detection. All models are trained on \texttt{train2017}, including $57K$ images and $150K$ person instances labeled with 17 keypoints. These models are evaluated on \texttt{val2017} containing 5000 images by using the mean average precision (AP) over 10 object key point similarity (OKS) thresholds as the metric.

\noindent \textbf{Implementation Details:} We evaluate DY-ReLU on two backbone networks (MobileNetV2 and MobileNetV3) and one head network used in \cite{Chen2019DynamicCA}. The head simply uses upsampling and four MobileNetV2's inverted residual bottleneck blocks. We compare DY-ReLU with its static counterpart in both \textit{backbone} and \textit{head}. The spatial and channel-wise DY-ReLU-C is used here, as we show that the spatial attention is important for keypoint detection, especially in the head network (see section \ref{sec:dy-relu-var-ablation}). Note that when using MobileNetV3 as backbone, we remove Squeeze-and-Excitation and replace either ReLU or h-swish by DY-ReLU. The number of linear functions in DY-ReLU is set as $K=2$. The initialization values of slope and intercept are set as $\alpha^1=1$, $\alpha^2=\beta^1=\beta^2=0$. The range of slope and intercept are set as $\lambda_a=1$ and $\lambda_b=0.5$, respectively. 

\noindent \textbf{Training setup:} We follow the training setup in \cite{sun2019deep}. All models are trained from scratch for 210 epochs, using Adam optimizer \cite{kingma:adam}. The initial learning rate is set as 1e-3 and is dropped to 1e-4 and 1e-5 at the $170^{th}$ and $200^{th}$ epoch, respectively. All human detection boxes are cropped from the image and resized to $256\times192$. The data augmentation includes random rotation ($[-\ang{45}, \ang{45}]$), random scale ($[0.65, 1.35]$), flipping, and half body data augmentation.

\noindent \textbf{Testing:} We use the person detectors provided by \cite{xiao2018simplebaseline} and follow the evaluation procedure in \cite{xiao2018simplebaseline,sun2019deep}. The keypoints are predicted on the average heatmap of the original and flipped images. The highest heat value location is then adjusted by a quarter offset from the highest response to the second highest response.

\begin{table*}[t]
	\begin{center}
		\footnotesize
		\begin{tabular}{c|r r r| l c c c c}
			\specialrule{.1em}{.05em}{.05em} 
			Backbone & Activation & Param & MAdds & AP &	AP$^{0.5}$ & AP$^{0.75}$ & AP$^M$ & AP$^L$\\
			\specialrule{.1em}{.05em}{.05em} 
			\multirow{2}{*}{MBNetV2 $\times1.0$}& ReLU & 3.4M & 993.7M  &64.6           &87.0  & 72.4 & 61.3 & 71.0 \\
			&DY-ReLU& 9.0M & 1026.9M & 68.1$_{(3.5)}$ &	88.5 & 76.2 & 64.8 & 74.3 \\
			\hline
			\multirow{2}{*}{MBNetV2 $\times0.5$}& ReLU & 1.9M & 794.8M & {59.3}         & {84.3} & 	{66.4} &	{56.2}&		{65.0}\\
			&DY-ReLU& 4.6M & 820.3M & 63.3$_{(4.0)}$ & 	86.3 &	71.4&		60.3&	69.2\\
			
			\hline
			\multirow{2}{*}{MBNetV3 Large} & ReLU/SE/HS       &  4.1M & 896.4M & 65.7          &87.4  & 74.1 & 62.3 & 72.2 \\
			&DY-ReLU&  10.1M & 926.6M & 67.2$_{(1.5)}$ & 88.2	 & 75.4 & 64.1 & 73.2 \\
			\hline
			\multirow{2}{*}{MBNetV3 Small}&  ReLU/SE/HS    & 2.1M & 726.9M & 57.1 & 83.8 & 63.7 &	55.0&		62.2\\
			&DY-ReLU& 4.8M & 747.9M & 60.7$_{(3.6)}$ & 85.7	 &	68.1 &	58.1	&	66.3\\
			\specialrule{.1em}{.05em}{.05em} 
		\end{tabular}
	\end{center}
	\caption{Comparing DY-ReLU with baseline activation functions (ReLU, SE or h-swish, denoted as HS) on COCO Keypoint detection. The evaluation is on validation set. The head structure in \cite{Chen2019DynamicCA} is used. DY-ReLU-C with $K=2$ is used in both backbone and head. 
	Note that SE blocks are removed when using DY-ReLU in MobileNetV3. The numbers in brackets denote the performance improvement over the baseline. DY-ReLU outperforms its static counterpart by a clear margin.
	}
	\label{table:coco-kp}
\end{table*}

\noindent \textbf{Main Results}: Table \ref{table:coco-kp} shows the comparison between DY-ReLU and its static counterpart in four different backbone networks (MobileNetV2 $\times 0.5$ and $\times 1.0$, MobileNetV3 Small and Large). The head network \cite{Chen2019DynamicCA} is shared for these four experiments. DY-ReLU outperforms baselines by a clear margin. It gains 3.5 and 4.0 AP when using MobileNetV2 with width multipler $\times 1.0$ and $\times 0.5$, respectively. It also gains 1.5 and 3.6 AP when using MobileNetV3-Large and MobileNetV3-Small, respectively. These results demonstrate that our method is also effective on keypoint detection.

\section{Conclusion}
In this paper, we introduce dynamic ReLU (DY-ReLU), which adapts a piecewise linear activation function dynamically for each input. Compared to its static counterpart (ReLU and its generalizations), DY-ReLU significantly improves the representation capability with negligible extra computation cost, thus is more friendly to efficient CNNs. Our dynamic ReLU can be easily integrated into existing CNN architectures. By simply replacing ReLU (or h-swish) in ResNet and MobileNet (V2 and V3) with DY-ReLU, we achieve solid improvement for both image classification and human pose estimation. We hope DY-ReLU becomes a useful component for efficient network architecture.



\clearpage
%
%
\bibliographystyle{splncs04}
\bibliography{egbib}
\end{document}